\documentclass{article}

\usepackage[final]{corl_2026}

\usepackage{graphicx}
\usepackage{amsmath}
\usepackage{amssymb}
\usepackage{booktabs}
\usepackage{multirow}
\usepackage[table]{xcolor}
\usepackage{colortbl}
\usepackage{wrapfig}
\usepackage{enumitem}
\usepackage{orcidlink}
\usepackage{listings}
\usepackage{caption}
\usepackage{subcaption}

\usepackage{hyperref}

\makeatletter
\renewcommand{\@noticestring}{%
  Submitted to the \@conferenceordinal\/ Conference on Robot Learning (CoRL \@conferenceyear).\par
  \textbf{Project page: }
  \href{https://zhangxd-666.github.io/UniTacVLA/}{\texttt{github.io/UniTacVLA}}%
}
\makeatother

\title{UniTacVLA: Unified Tactile Understanding and Prediction in Vision Language Action Models}

\author{
{\normalfont
\begin{tabular}{c}
Xidong Zhang$^{1,2,6,*}$ \quad
Yichi Zhang$^{1,6,*}$ \quad
Jiaxin Shi$^{3,6}$ \quad
Fucai Zhu$^{2,6}$ \\
Siyu Zhu$^{4}$ \quad
Michael Yu Wang$^{2,6}$ \quad
Xiaojun Wu$^{1}$ \quad
Weihao Yuan$^{5,6}$ \\
\\[-0.5ex]
{\small $^{1}$Harbin Institute of Technology \quad
$^{2}$Great Bay University \quad
$^{3}$Shanghai Jiao Tong University} \\
{\small $^{4}$Fudan University \quad
$^{5}$Nanjing University \quad
$^{6}$Daimon Robotics} \\
\\[-0.5ex]
{\small $^{*}$Equal contribution.}
\end{tabular}
}
}

\begin{document}
\maketitle

\selectfont

\vspace{-8mm}
\begin{abstract}
Vision-language-action (VLA) models have achieved strong performance in many robotic manipulation tasks, yet remain limited in contact-rich dexterous manipulation. To overcome this limitation, recent vision-tactile-language-action (VTLA) methods incorporate tactile sensing into VLA models to provide direct contact information. However, they typically treat tactile signals as passive auxiliary inputs, making it difficult to model tactile semantics and future physical interactions.
To this end, we propose a unified tactile learning framework for contact-rich manipulation that models tactile signals as dynamic interaction cues for both contact understanding and prediction. Specifically, we construct a unified tactile latent space and jointly model current tactile states and future contact changes through tactile chain-of-thought reasoning and coarse-to-fine future tactile prediction, thereby forming a state-aware and dynamics-aware tactile prior. Based on this prior, we introduce an action-tactile mixed controller that combines real-time and predicted tactile feedback to refine low-frequency action chunks with high-frequency corrections.
Real-world experiments on four categories of contact-rich tasks, including adjustment, wiping, insertion, and assembly, under both clean and perturbed settings, show that our method improves success rate, manipulation accuracy, and contact robustness over existing methods, demonstrating its effectiveness in dexterous physical interaction.
\end{abstract}

\vspace{-3mm}

\vspace{-3mm}
\section{Introduction}
\label{sec:introduction}
\vspace{-3mm}
Current vision-language-action (VLA) models have achieved strong performance in routine robotic manipulation, but remain limited in contact-rich dexterous tasks. This limitation stems from two key factors. First, vision alone cannot reliably capture local and transient contact states, such as contact onset, slipping, jamming, and subtle alignment errors, especially under occlusion. Second, contact-rich manipulation requires continuous real-time adjustment to evolving contact conditions, rather than executing a single upfront plan, whereas most VLA systems rely on low-frequency, open-loop action chunks. As a result, small uncorrected deviations can quickly accumulate into collision, slipping, or jamming, leading to task failure.
Incorporating tactile signals into VLA models is a natural solution, as touch directly captures local physical interactions that are difficult to observe visually. However, most tactile-enhanced VLA methods still treat touch as a passive auxiliary modality, with limited explicit modeling of tactile interaction semantics and dynamics~\citep{see_hear,tactile_vla,tac_vla,vtla,play_to_the_score}. Although recent tactile prediction methods~\citep{pred1,pred2,dreamtac_vla,omnivta} forecast future tactile states, they are often formulated as standalone dynamics models or world-model rollouts, making it difficult to translate tactile prediction into tactile reasoning, action generation, or online correction.

\begin{figure}[t]
\vspace{-5mm}
    \centering
    \includegraphics[width=\linewidth]{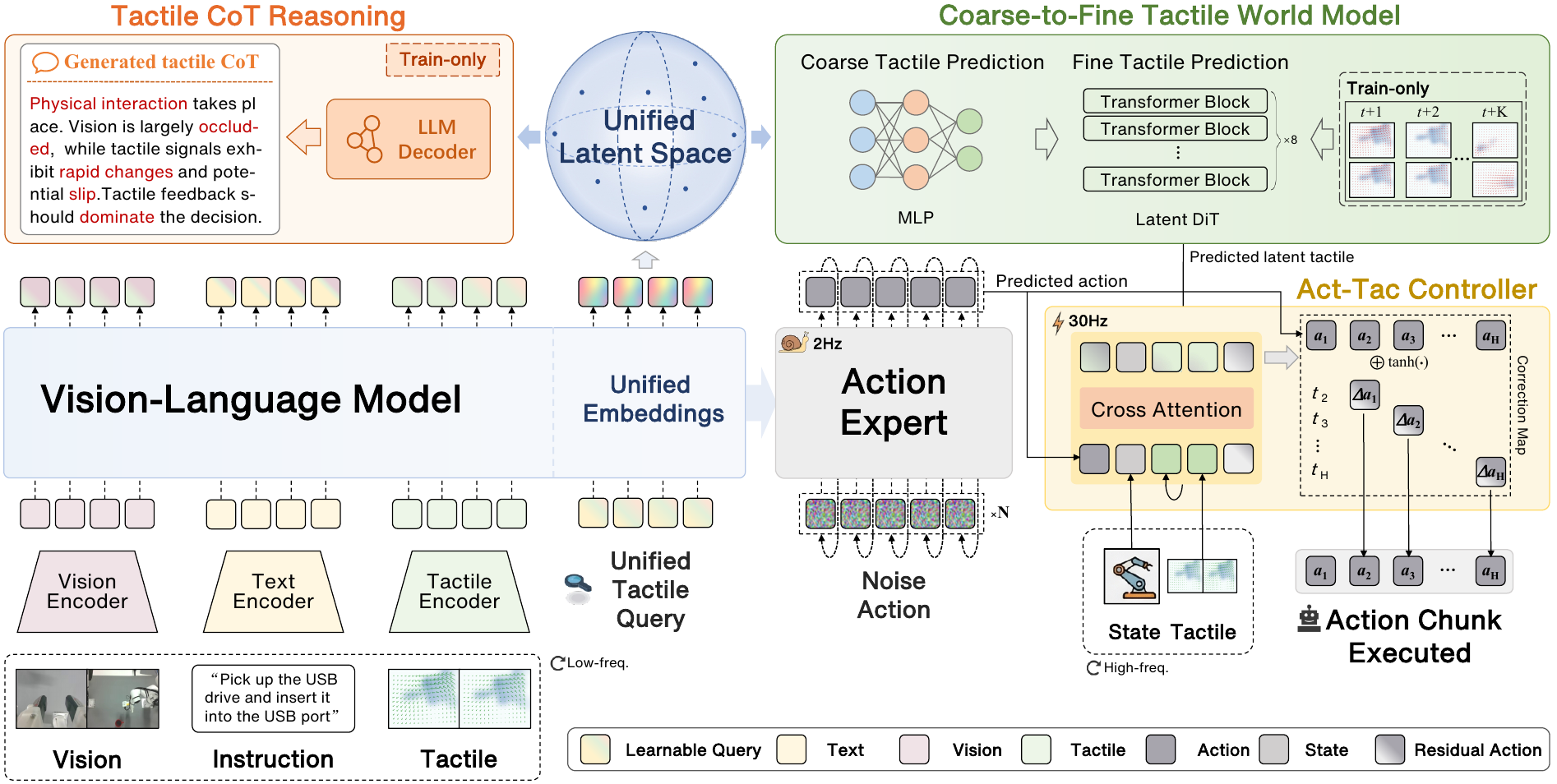}
    \vspace{-3mm}
    \caption{Overview of UniTacVLA. UniTacVLA learns a unified tactile latent space for contact-rich manipulation by combining tactile chain-of-thought reasoning with coarse-to-fine future tactile prediction. The learned proactive tactile prior captures both current contact states and future tactile dynamics. The predicted tactile dynamics are further used by the action-tactile mixed controller to refine low-frequency action chunks with high-frequency tactile-guided corrections, enabling more robust contact-rich manipulation.}
    \vspace{-4mm}
\label{fig:overview}
\end{figure}

To address these limitations, we propose UniTacVLA, a unified tactile learning framework that integrates both tactile understanding and future tactile prediction for contact-rich manipulation, as illustrated in Fig.~\ref{fig:overview}. Instead of passively fusing tactile inputs with vision and language, UniTacVLA learns a compact tactile latent space that is both state-aware and dynamics-aware, capturing current contact conditions while anticipating future contact evolution.
Specifically, we introduce learnable unified tactile tokens into the VLM to extract contact-related information from multimodal observations and construct a unified tactile latent space. To make this space state-aware, we apply tactile chain-of-thought (T-CoT) supervision, encouraging the model to reason about contact stages, contact conditions, and potential failure modes. To make it dynamics-aware, we further design a coarse-to-fine tactile prediction objective that models the evolution of tactile latent states during action execution. By coupling current-state reasoning with future contact prediction, UniTacVLA forms a proactive tactile prior for downstream control. Finally, we introduce an action-tactile mixed controller that refines low-frequency action chunks via high-frequency corrections conditioned on both real-time and predicted tactile feedback.

We evaluate UniTacVLA through extensive real-world experiments on eight contact-rich manipulation tasks across four task categories, covering both clean settings without disturbances and perturbed settings with external disturbances. Experimental results show that UniTacVLA consistently improves success rate and manipulation robustness over existing methods, achieving state-of-the-art performance under both settings. These results demonstrate that unified tactile understanding and prediction provide an effective prior for robust contact-rich robotic manipulation and correction.

Our contributions are summarized as follows:
\vspace{-2mm}
\begin{itemize} [leftmargin=*]
    \item We propose a unified tactile learning framework that models tactile signals as predictive interaction cues rather than passive auxiliary inputs.
    \item We introduce a unified tactile latent space supervised by tactile chain-of-thought reasoning and coarse-to-fine future tactile prediction, enabling joint modeling of current tactile states and future contact dynamics.
    \item We develop an action-tactile mixed controller that uses proactive tactile priors and online tactile feedback to refine low-frequency action chunks with high-frequency corrections.
\end{itemize}

\vspace{-3mm}
\section{Related Work}
\label{sec:related_work}

\vspace{-3mm}
\textbf{Tactile fusion for robotic manipulation.}
Tactile sensing provides contact-level feedback that is difficult to infer from vision alone and is crucial for contact-rich manipulation~\cite{calandra2018more,dong2021tactile,qi2023general,sunil2023visuotactile,schoettler2020deep,forcemimic,rdp,huang20243d}. While general VLA models enable language-conditioned control, they mainly rely on RGB and proprioception, leaving contact states under-modeled~\cite{act,openvla,xvla,rdt,pi0,pi05,focusvla}. Recent VTLA models incorporate tactile or force observations into VLA-style policies~\cite{hao2025tla,tactile_vla,tac_vla,vtla,vla_touch,forcevla,jones2025beyond}, but most treat tactile inputs as auxiliary tokens rather than explicit contact semantics for policy reasoning. In contrast, our method aligns tactile feedback with semantic reasoning and action generation, enabling contact-aware multimodal decision making.


\vspace{-2mm}
\textbf{Predictive modeling and tactile world models.}
Predicting task-relevant content or future world evolution helps policies focus on manipulation-critical context and incorporate world-model priors~\cite{reconvla,dreamvla,spatialforcing,dualcotvla,vipra,kim2026cosmos,ha2018world,dreamerv3,assran2025v,maes2026leworldmodel}. In tactile manipulation, tactile prediction and tactile world models have been explored to capture contact dynamics~\cite{pred1,pred2,robopack,dreamtac_vla,omnivta}. However, existing methods typically focus on either semantic reasoning or physical dynamics modeling. In contrast, our method jointly learns tactile semantic reasoning and future tactile prediction: T-CoT captures task stages and contact states, while coarse-to-fine tactile prediction anticipates tactile evolution for contact-aware action generation.

\vspace{-2mm}
\textbf{Slow-fast control for contact-rich manipulation.}
To reduce the latency caused by open-loop action chunking, recent works adopt slow-fast hierarchical control that combines low-frequency reasoning with high-frequency feedback control~\cite{fast_in_slow,last0,rdp,favla,forcevla,forcevla2}. However, most existing methods perform reactive correction based only on current tactile or force feedback, without anticipating future contact failures. Our method preserves online feedback control while using predicted future tactile states as a prior, enabling proactive action refinement before contact failure occurs.

\vspace{-2mm}
\section{Method}
\label{sec:method}
\vspace{-3mm}

We introduce \textbf{UniTacVLA}, a unified tactile understanding and future tactile prediction paradigm for VTLA policies. It uses unified tactile tokens as a bridge between tactile understanding and prediction, thereby constructing predictive tactile priors that improve the robot's ability to model physical interaction processes. Based on predicted future tactile latents and real-time tactile feedback, the action-tactile mixed controller progressively refines action execution and enables high-frequency closed-loop control for more stable fine-grained manipulation.
\vspace{-2mm}
\subsection{Preliminaries}
\label{sec:method:preliminaries}
\vspace{-2mm}

\textbf{VTLA architecture.}
Given visual observation $V_t$, language instruction $L$, and tactile observation $T_t$, a VTLA policy predicts an action chunk:
\[
A_{t:t+H} = \pi_\phi(\sigma_\theta^{\mathrm{VLM}}(V_t, L, T_t)).
\]
Here, $\sigma_\theta^{\mathrm{VLM}}$ denotes the VLM encoder parameterized by $\theta$, $\pi_\phi$ is the action policy parameterized by $\phi$, and $A_{t:t+H}=\{a_t,\ldots,a_{t+H}\}$ represents the predicted action chunk over horizon $H$.

\textbf{Visuo-tactile encoder.}
We use a variational masked autoencoder (VMAE) to obtain compact tactile representations from tactile observations~\citep{vmae}. At time step $t$, tactile observations from the left and right grippers are concatenated as $X_t \in \mathbb{R}^{H_T \times W_T \times 3 \times 2}$, where the three channels represent deformation of the sensor surface along the $z$, $x$, and $y$ axes. A Transformer encoder compresses $X_t$ into tactile tokens:
\[
z_t^{\mathrm{tac}} = \sigma_\theta^{\mathrm{Enc}}(X_t), \quad z_t^{\mathrm{tac}} \in \mathbb{R}^{N \times d},
\]
where $N$ and $d$ denote the number and dimension of tactile tokens. A Transformer decoder reconstructs the tactile observation,
\[
\hat{X}_t = \sigma_\theta^{\mathrm{Dec}}(z_t^{\mathrm{tac}}),
\]
which encourages the tactile tokens to preserve structured physical interaction information. 
\vspace{-2mm}
\subsection{Unified Tactile Understanding and Prediction}
\label{sec:method:uup}
\vspace{-2mm}
Unlike prior methods that treat tactile sensing merely as a passive input, our unified understanding and prediction paradigm enables the model to actively learn, interpret, and forecast tactile information. Moreover, we jointly couple the learned prior with action generation and correction.

\textbf{Unified tactile tokens.}
To precisely extract task-relevant tactile information, we introduce a set of unified tactile tokens $Q_t$, a set of $N$ learnable query embeddings. Within the VLM $\sigma_\theta^{\mathrm{VLM}}$, $Q_t$ extract task-relevant tactile information from multimodal inputs:
\begin{equation}
z_t^{(v,l,t,q)} = \sigma_\theta^{\mathrm{VLM}}(V_t, L, T_t, Q_t),
\end{equation}
where $V_t$ is the visual observation at time step $t$, $L$ is the language instruction, $T_t$ represents real tactile observations, and $z_t^{(v,l,t,q)}$ denotes the policy-conditioning latent state. In the policy module, the tactile feature $z_t^{t}$ and tactile prior $z_t^{q}$ are incorporated as part of the policy state $z_t$ to guide action generation:
\begin{equation}
A_{t:t+H} = \pi_{\phi}(z_t^{(v,l,t,q)}).
\end{equation}
In this way, unified tactile tokens aggregate tactile prior and naturally integrate tactile perception into action decision making.

\textbf{Tactile chain of thought.} 
To strengthen the semantic understanding of tactile states, we further introduce a T-CoT mechanism. Specifically, $z_t^q$ is used as a conditioning prefix to the pretrained large language model $\sigma_\theta^{\mathrm{LLM}}$, which generates the tactile reasoning chain in an auto-regressive manner:
\begin{equation}
p_\theta^{\mathrm{LLM}}(\mathrm{CoT}_t \mid z_t^q)
=
\prod_{\ell=1}^{L}
p_\theta^{\mathrm{LLM}}
\left(c_{t,\ell} \mid z_t^q, c_{t,<\ell}\right),
\end{equation}
where $c_{t,\ell}$ denotes the $\ell$-th token in the tactile reasoning chain.
The generated CoT text serves as a high-level semantic proxy for the current tactile state, describing tactile observations in natural language and providing cues about possible dynamics and manipulation. This mechanism encourages $z_t^q$ to function as a tactile information bottleneck that preserves task-relevant semantic information, thereby promoting more structured tactile representations during cross-modal alignment.

\textbf{Coarse-to-fine tactile prediction.}
Beyond learning task-relevant tactile semantics, UniTacVLA further predicts future tactile states. 
Since dense tactile signals are high-dimensional and fine-grained, direct prediction can slow joint optimization with action learning. We therefore adopt a coarse-to-fine scheme that decouples global trend prediction and local detail refinement, thereby accelerating convergence and improving prediction quality.

To capture the trend of tactile changes and establish a stable global prediction target, while avoiding the direct generation of high-dimensional tactile details from a compact semantic prior, we employ a lightweight multilayer perceptron (MLP) to map $z_t^q$ into a coarse future tactile representation:
\begin{equation}
z_t^{\mathrm{coarse}} = \sigma_\theta^{\mathrm{MLP}}(z_t^q).
\end{equation}

Then, the DiT conditions on $z_t^q$ and progressively injects local details and high-frequency information into the coarse representation, producing a more accurate future tactile prediction:
\begin{equation}
z_t^{\mathrm{fine}} = \sigma_\theta^{\mathrm{DiT}}(z_t^{\mathrm{coarse}}, z_t^q).
\end{equation}
For visualization purposes only, we can further use the frozen VMAE decoder to map the refined latent representation back to future tactile images:
\begin{equation}
\hat{T}_{t:t+H} = \sigma_\theta^{\mathrm{Dec}}(z_t^{\mathrm{fine}}).
\end{equation}
During training and inference, UniTacVLA directly uses the predicted tactile latent representations without decoding them into images. This coarse-to-fine prediction strategy preserves global consistency while refining local tactile details, providing more accurate foresight for robot manipulation.

\vspace{-2mm}
\subsection{Action-Tactile Mixed Controller}
\label{sec:method:controller}
\vspace{-2mm}
We introduce a high-frequency tactile controller that corrects robot actions using both predicted future tactile latents and real-time tactile feedback. By combining contact foresight with online feedback, the controller enables more robust action correction during dynamic interactions.

The controller is a lightweight Transformer that takes the current action $a_t$, the predicted future tactile latent $z_t^{\mathrm{tac_{pred}}}$, and the current real tactile observation $z_t^{\mathrm{tac_{curr}}}$ as input, and outputs a residual correction $\Delta a_t$:
\begin{equation}
\Delta a_t = \tanh\left(
\pi_{\theta}^{\mathrm{Ctrl}}(
a_t, z_t^{\mathrm{tac_{pred}}}, z_t^{\mathrm{tac_{curr}}}
)
\right).
\end{equation}
Here, $\tanh$ bounds the correction magnitude to prevent it from dominating the backbone action. The final executed action is obtained by adding the bounded correction to the final action:
\begin{equation}
a_t^{\mathrm{final}} = a_t + \Delta a_t.
\end{equation} 
Our controller enables the robot to recover from perturbations and erroneous states, thereby improving the success rate of contact-rich manipulation.

\begin{figure}[t]
\vspace{-5mm}
    \centering
    \includegraphics[width=\linewidth]{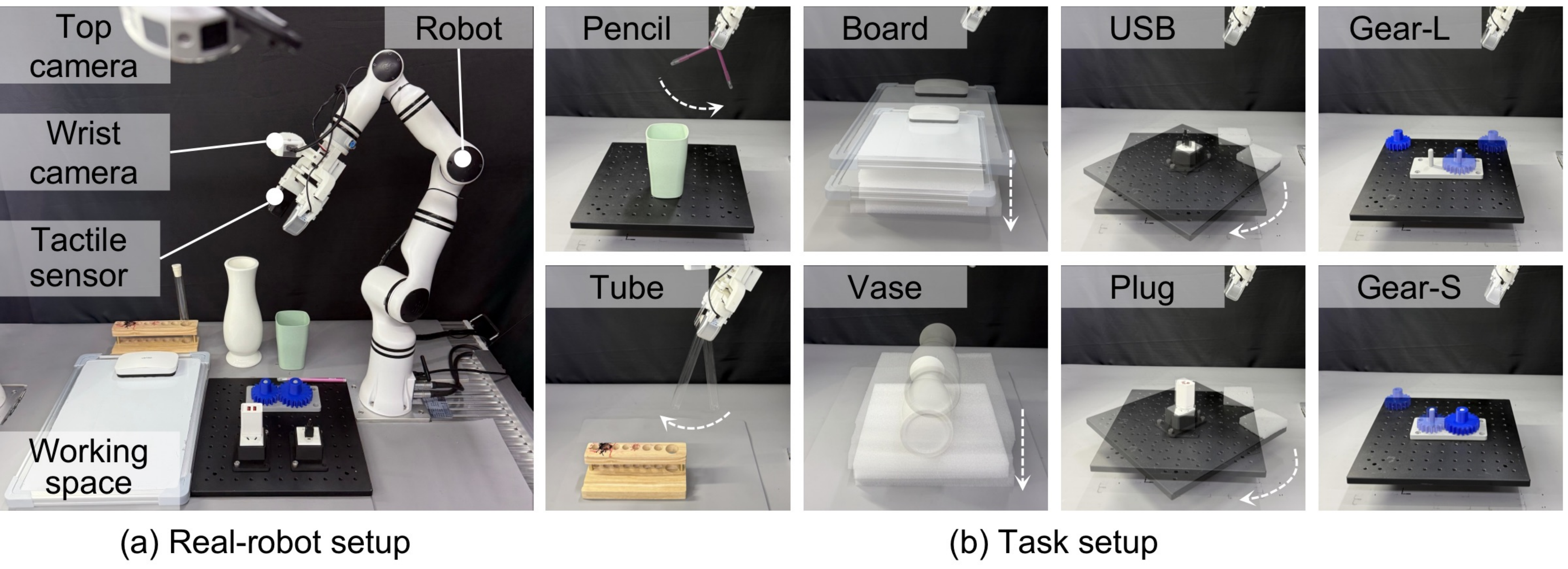}
    \vspace{-6mm}
    \caption{Overview of the real-robot setup and task setup. We evaluate UniTacVLA on four categories of contact-rich manipulation tasks covering eight subtasks in total, using a RealMan 7-DoF robotic arm equipped with a DM-Tac W visuo-tactile sensor. We further introduce human-induced disturbances, as illustrated in (b), to evaluate the robustness of different methods under perturbed contact conditions.}
    \vspace{-4mm}
    \label{fig:setup}
\end{figure}

\vspace{-2mm}
\subsection{Training}
\label{sec:method:training}
\vspace{-2mm}

\textbf{Tactile representation learning.}
We train the tactile encoder with a variational masked autoencoding objective, where the encoder parameterizes a latent distribution $q(z|x)$ from tactile observations $x$, and the decoder reconstructs tactile tokens from sampled latents, as
\begin{equation}
\mathcal{L}_{\mathrm{VMAE}}
=
(1-\alpha)\mathcal{L}_{\mathrm{mask}}
+
\alpha \mathcal{L}_{\mathrm{vis}}
+
\beta
D_{\mathrm{KL}}
\left(
q(z|x) , \mathcal{N}(0,I)
\right),
\end{equation}
where $\mathcal{L}_{\mathrm{mask}}$ and $\mathcal{L}_{\mathrm{vis}}$ reconstruct masked and visible tactile tokens, respectively, $\alpha, \beta$ denote two weighting factors, and $D_{\mathrm{KL}}$ denotes the KL divergence to regularize the variational latent space toward a standard Gaussian prior.

\textbf{Two-stage VTLA training.}
Our method is trained in two stages. In the first stage, we jointly optimize the backbone policy with action prediction, semantic prediction, and tactile prediction objectives:
\begin{equation}
\mathcal{L}_{\mathrm{stage1}}
=
\mathcal{L}_{\mathrm{FM}}^{\mathrm{action}}
+
\lambda_{\mathrm{sem}}\mathcal{L}_{\mathrm{CE}}^{\mathrm{semantic}}
+
\lambda_{\mathrm{coarse}}\mathcal{L}_{1}^{\mathrm{coarse}}
+
\lambda_{\mathrm{fine}}\mathcal{L}_{\mathrm{FM}}^{\mathrm{fine}}.
\end{equation}
Here, $\mathcal{L}_{\mathrm{FM}}^{\mathrm{action}}$ denotes the flow-matching loss for action generation, $\mathcal{L}_{\mathrm{CE}}^{\mathrm{semantic}}$ is the cross-entropy loss for semantic prediction, and $\mathcal{L}_{1}^{\mathrm{coarse}}$ and $\mathcal{L}_{FM}^{\mathrm{fine}}$ supervise coarse- and fine-level tactile prediction, respectively. The weighting factors $\lambda_{\mathrm{sem}}$, $\lambda_{\mathrm{coarse}}$, and $\lambda_{\mathrm{fine}}$ balance the contributions of the semantic and tactile auxiliary objectives. The first stage is trained exclusively on clean expert demonstrations without any human-induced disturbances, providing reliable supervision for tactile understanding, future tactile prediction, and action generation.
In the second stage, we train the high-frequency tactile controller for residual action correction:
\begin{equation}
\mathcal{L}_{\mathrm{stage2}}
=
\mathcal{L}_1^{\mathrm{ctrl}}.
\end{equation}
Here, $\mathcal{L}_1^{\mathrm{ctrl}}$ is the $\ell_1$ loss on the predicted residual action. The controller is trained on a mixture of clean expert demonstrations and trajectories with human-induced disturbances and recoveries from erroneous states. This composition exposes the controller to corrective behaviors while preserving the action distribution learned in the first stage.



\vspace{-2mm}
\section{Experiments}
\label{sec:experiments}
\vspace{-2mm}
We conduct extensive real-world experiments to evaluate the effectiveness of UniTacVLA on 8 contact-rich manipulation tasks. The experiments are designed to answer the following key questions:
\textbf{Q1:} Can UniTacVLA provide meaningful stage-level understanding and coordinate different modalities during manipulation?
\textbf{Q2:} Can UniTacVLA accurately predict future tactile states?
\textbf{Q3:} Can the unified tactile learning effectively guide action generation and improve policy performance?
\textbf{Q4:} Does the high-frequency correction network provide beneficial residual corrections to the generated actions?
\textbf{Q5:} Can our method still improve performance when deployed on robots without tactile sensors at inference time?

\vspace{-2mm}
\subsection{Experimental Setup}
\vspace{-2mm}

Experiments are conducted on a RealMan 7-DoF robotic arm equipped with a 1-DoF gripper and two fingertip-mounted DM-Tac W visuo-tactile sensors. We evaluate UniTacVLA on four categories of contact-rich tasks, including pose adjustment, insertion, wiping, and gear assembly, covering eight subtasks in total. To assess robustness, we test each method under both clean and perturbed settings, where mild human-induced disturbances, including human poking, platform orientation changes, and surface height variations, are introduced during execution. Success rate within a fixed time limit is used as the primary metric, with 50 trials conducted for each setting. We compare UniTacVLA with $\pi_0$~\cite{pi0}, $\pi_{0.5}$~\cite{pi05}, and two visuo-tactile baselines reproduced based on $\pi_{0.5}$, VTLA~\cite{vtla} and TacVLA~\cite{tac_vla}. Specifically, VTLA extends $\pi_{0.5}$ by incorporating visuo-tactile sensory input, while TacVLA further integrates an action-tactile controller to enhance action refinement during contact-rich interactions. All methods are trained on four NVIDIA A100 GPUs and deployed on an NVIDIA RTX 5090 GPU. More implementation details are provided in the Appendix.

\definecolor{oursgray}{gray}{0.92}
\begin{table}[b]
\vspace{-3mm}
\centering
\scriptsize
\setlength{\tabcolsep}{3pt}
\begin{tabular}{l*{16}{c}}
    \toprule
    \textbf{Task}
    & \multicolumn{4}{c}{\textbf{Adjust}}
    & \multicolumn{4}{c}{\textbf{Wipe}}
    & \multicolumn{4}{c}{\textbf{Insert}}
    & \multicolumn{4}{c}{\textbf{Assemble}} \\
    \cmidrule(lr){2-5}\cmidrule(lr){6-9}\cmidrule(lr){10-13}\cmidrule(lr){14-17}

    \textbf{Subtask}
    & \multicolumn{2}{c}{\textbf{pencil}} & \multicolumn{2}{c}{\textbf{tube}}
    & \multicolumn{2}{c}{\textbf{board}} & \multicolumn{2}{c}{\textbf{vase}}
    & \multicolumn{2}{c}{\textbf{usb}} & \multicolumn{2}{c}{\textbf{plug}}
    & \multicolumn{2}{c}{\textbf{gearL}} & \multicolumn{2}{c}{\textbf{gearS}} \\
    \cmidrule(lr){2-3}\cmidrule(lr){4-5}\cmidrule(lr){6-7}\cmidrule(lr){8-9}
    \cmidrule(lr){10-11}\cmidrule(lr){12-13}\cmidrule(lr){14-15}\cmidrule(lr){16-17}

    \textbf{Method}
    & cln & ptb & cln & ptb & cln & ptb & cln & ptb
    & cln & ptb & cln & ptb & cln & ptb & cln & ptb \\
    \midrule

    $\pi_0$~\cite{pi0}              
    & 42 & 14 & 34 & 10 & 38 & 0  & 24 & 0
    & 18 & 4  & 22 & 0  & 4  & 0  & 8  & 0  \\

    $\pi_{0.5}$~\cite{pi05}          
    & 50 & 18 & 40 & 10 & 40 & 0  & 22 & 0
    & 20 & 8  & 18 & 10 & 8  & 0  & 10 & 0  \\

    $\pi_{0.5}$-VTLA~\cite{vtla}  
    & 72 & 36 & 64 & 20 & 44 & 6  & 48 & 0
    & 22 & 14 & 18 & 10 & 18 & 8  & 12 & 4  \\
    
    $\pi_{0.5}$-TacVLA~\cite{tac_vla}  
    & 86 & 40 & 70 & 28 & 52 & 8  & 56 & 4
    & 30 & 22 & 28 & 18 & 22 & 4  & 18 & 6  \\

    \rowcolor{oursgray}
    UniTacVLA${^\dagger}$ (ours)   
    & \textbf{88} & 50 & 60 & 32 & 56 & 6  & 50 & 6
    & 42 & 18 & 44 & 24 & 24 & 0  & 20 & 0  \\

    \rowcolor{oursgray}
    \textbf{UniTacVLA (ours)}            
    & \textbf{88} & \textbf{86} & \textbf{74} & \textbf{64}
    & \textbf{76} & \textbf{60} & \textbf{64} & \textbf{54}
    & \textbf{62} & \textbf{58} & \textbf{54} & \textbf{48}
    & \textbf{44} & \textbf{26} & \textbf{50} & \textbf{32} \\
    
    \bottomrule
\end{tabular}
\vspace{4pt}
\caption{
Comparative results on contact-rich manipulation tasks. For each subtask, \textit{cln} 
(clean, without perturbation) and \textit{ptb} (perturbed, with human-induced disturbance) 
report success rates over 50 trials each. ${\dagger}$ denotes our method without real tactile input at inference time.
}
\vspace{-3mm}
\label{tab:comparison}
\end{table}

\begin{figure}[t]
\vspace{-5mm}
    \centering
    \includegraphics[width=\linewidth]{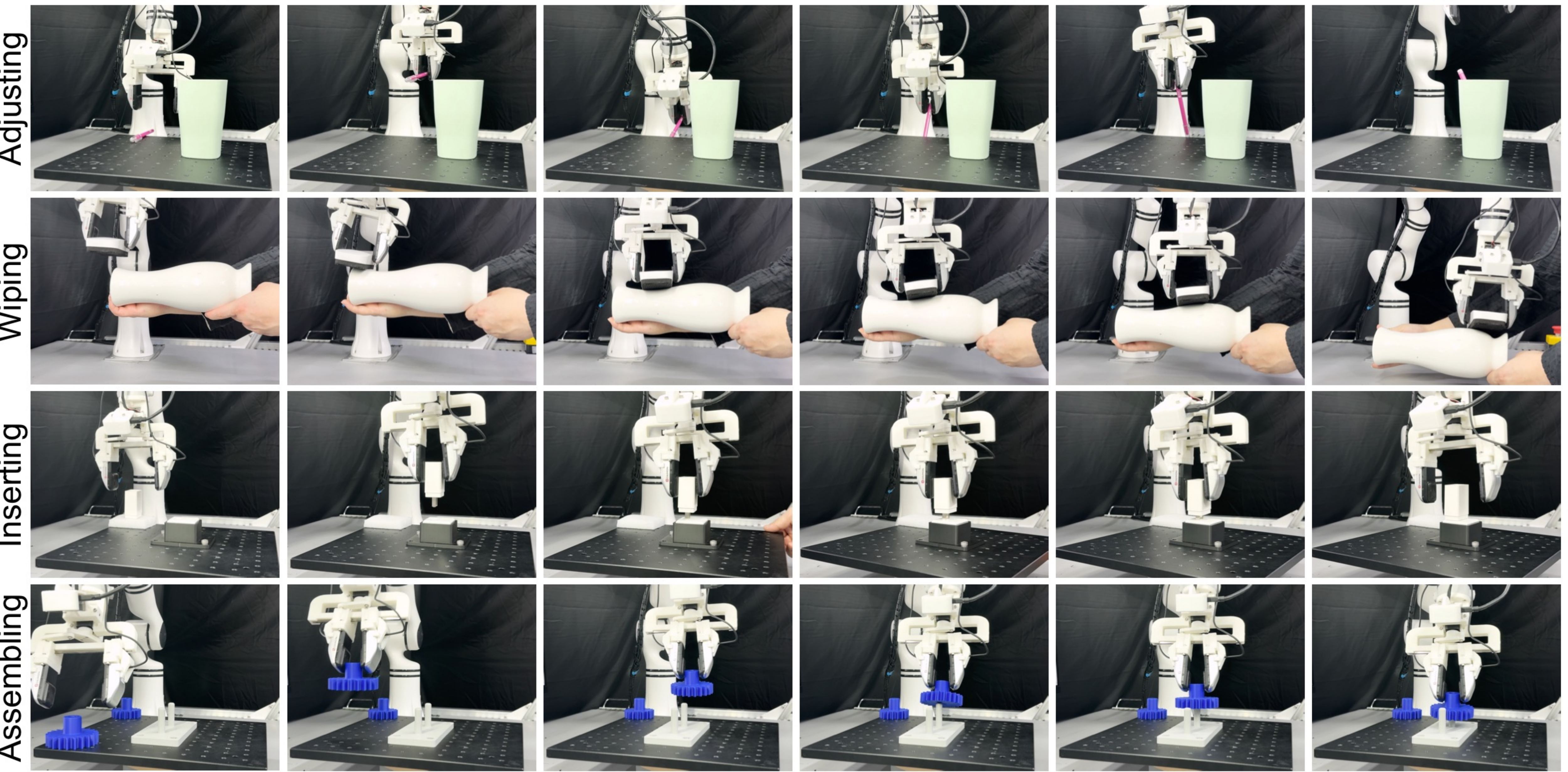}
    \vspace{-3mm}
    \caption{Demonstration of the four contact-rich manipulation subtasks evaluated in our experiments.}
    \label{fig:series}
    \vspace{-4mm}
\end{figure}

\vspace{-2mm}
\subsection{Comparative Experiments}
\label{sec:experiments:comparison}
\vspace{-2mm}

As shown in Table~\ref{tab:comparison}, our method achieves the highest average success rate under both clean and perturbed settings. 
The success of these tasks depends not only on visual localization, but also on accurate contact regulation: adjustment tasks require the policy to detect perturbation-induced changes in contact modes and respond properly; wiping tasks require maintaining appropriate contact force during continuous motion; and insertion and assembly tasks require tactile feedback for highly dynamic fine alignment under occlusion.
Although $\pi_{0.5}$-VTLA and $\pi_{0.5}$-TacVLA also use tactile observations, they simply concatenate tactile signals as ordinary tokens and therefore yield only limited improvement, indicating that naive tactile fusion is insufficient for contact-aware action reasoning. In contrast, our UniTacVLA explicitly models tactile understanding and future tactile prediction, enabling the policy to better infer contact stages and adjust actions accordingly~\textbf{(Q3)}. Moreover, the competitive performance achieved without real tactile input at inference time indicates that training with our unified tactile supervision helps the model acquire useful contact priors, thereby improving action generation even when tactile observations are unavailable~\textbf{(Q5)}.

\vspace{-3mm}
\subsection{Ablation Studies}
\label{sec:experiments:ablation}
\vspace{-2mm}

\begin{figure}[t]
\centering
\vspace{-5mm}
\begin{minipage}{0.65\linewidth}
\centering
\resizebox{\linewidth}{!}{
\begin{tabular}{ccccc|c}
\toprule
Tactile & T-CoT & Coarse pred & Fine pred & Controller & SR (\%) \\
\midrule
$\times$ & $\times$ & $\times$ & $\times$ & $\times$ & 18 \\
$\checkmark$ & $\times$ & $\times$ & $\times$ & $\times$ & 30 \\
$\checkmark$ & $\checkmark$ & $\times$ & $\times$ & $\times$ & 36 \\
$\checkmark$ & $\checkmark$ & $\checkmark$ & $\times$ & $\times$ & 44 \\
$\checkmark$ & $\checkmark$ & $\checkmark$ & $\checkmark$ & $\times$ & 52 \\
$\checkmark$ & $\checkmark$ & $\checkmark$ & $\checkmark$ & $\checkmark$ & 62 \\
\bottomrule
\end{tabular}
}
\vspace{-1mm}
\captionof{table}{Ablation study of different components on the USB task without disturbance.}
\label{tab:ablation}
\end{minipage}
\hfill
\begin{minipage}{0.3\linewidth}
\centering
\includegraphics[width=\linewidth]{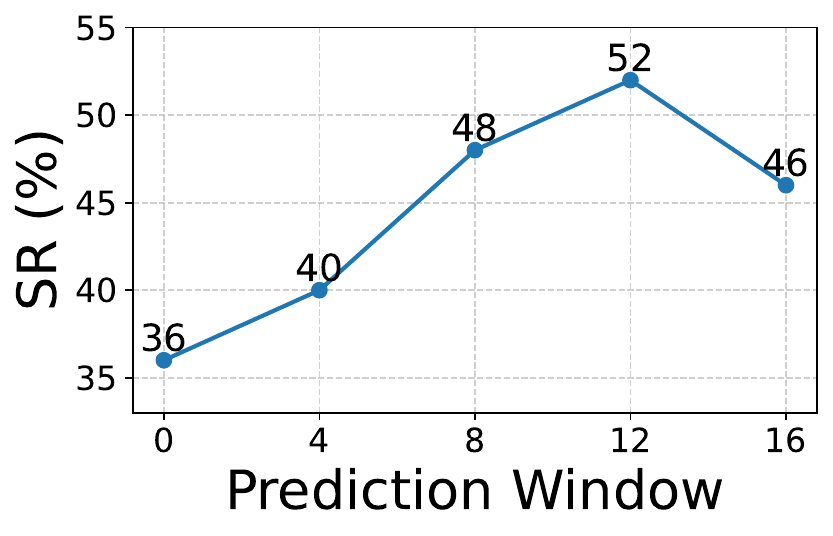}
\vspace{-3mm}
\captionof{figure}{Effect of prediction window size on USB task without disturbance.
}
\label{fig:prediction_window_sr}
\end{minipage}
\vspace{-4mm}
\end{figure}

\begin{figure}[t]
\vspace{-1mm}
    \centering
    \includegraphics[width=\linewidth]{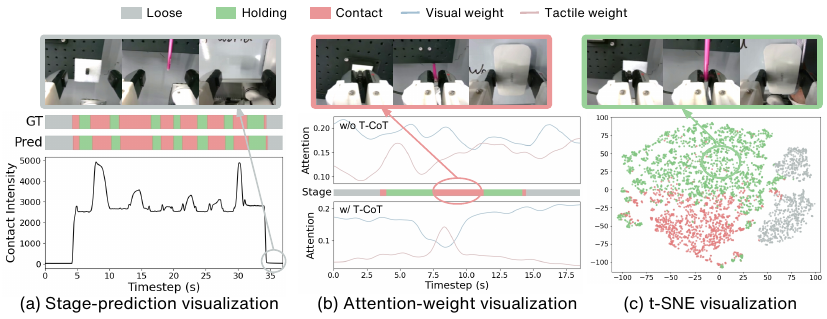}
    \vspace{-4mm}
    \caption{Qualitative results of T-CoT reasoning. (a) Our method accurately predicts rapid contact-stage transitions in board wiping despite subtle visual changes. (b) In USB insertion, T-CoT guides the model to place greater attention on tactile inputs during strong contact. (c) In the pencil adjustment task, unified tactile tokens form contact-aware semantic clusters under different contact patterns.}
    \vspace{-5mm}
    \label{fig:cot}
\end{figure}

%
\textbf{1) Tactile chain of thought (T-CoT).}
We first investigate the contribution of T-CoT in Table~\ref{tab:ablation}. Compared to the variant with tactile input alone, introducing T-CoT improves the success rate from 30\% to 36\%, indicating that explicitly reasoning about contact stages provides useful intermediate guidance for action prediction. To further understand this effect, we visualize the behavior of T-CoT in Fig.~\ref{fig:cot}. As shown in Fig.~\ref{fig:cot}(a), during the board-wiping task, the visual change around contact transitions is subtle, while the contact intensity varies significantly. Moreover, the rapid alternation between holding and contact makes contact-stage recognition challenging. Nevertheless, our method accurately captures these fine-grained contact changes, demonstrating its sensitivity to tactile interaction dynamics. Beyond contact-stage prediction, T-CoT also influences multimodal decision making. In Fig.~\ref{fig:cot}(b), we analyze the attention weights in the USB insertion task. When strong contact occurs, the model assigns higher attention to the tactile modality, suggesting that the generated reasoning not only describes the current interaction stage but also guides the model to focus on the most informative modality. Finally, Fig.~\ref{fig:cot}(c) shows the distribution of unified tactile tokens in the pencil adjustment task. Different contact patterns form distinguishable semantic clusters, further indicating that our method learns contact-aware tactile representations~\textbf{(Q1)}.

\begin{figure}[t]
\vspace{-5mm}
    \centering
    \includegraphics[width=\linewidth]{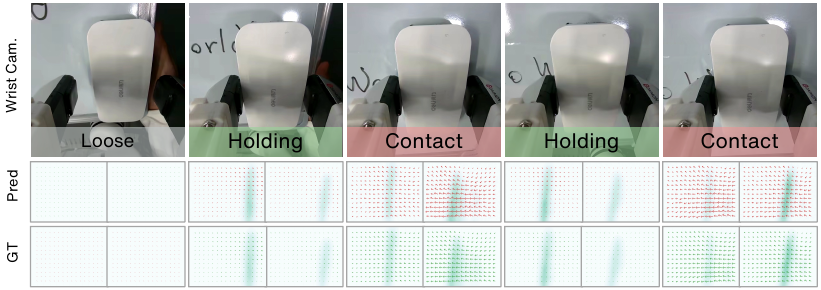}
    \vspace{-5mm}
    \caption{Qualitative results of coarse-to-fine future tactile prediction on the board-wiping task with disturbance.}
    \label{fig:predict}
\end{figure}

\textbf{2) Coarse-to-fine tactile world model.}
We further evaluate the contribution of the coarse-to-fine tactile world model. As shown in Table~\ref{tab:ablation}, adding coarse tactile prediction improves the success rate from 36\% to 44\%, while further introducing fine tactile prediction increases it to 52\%. This validates the necessity of the coarse-to-fine design: coarse prediction captures high-level contact evolution, whereas fine prediction refines local tactile dynamics for more accurate action prediction. We further visualize the predicted tactile signals in Fig.~\ref{fig:predict}. Even when the tactile stage changes rapidly, our method consistently captures the temporal dynamics of tactile signals, providing future-aware tactile cues for the manipulation~\textbf{(Q2)}. Finally, Fig.~\ref{fig:prediction_window_sr} studies the effect of the prediction window size. A short window provides insufficient future tactile context for effective controller interaction, while an overly long window requires predicting distant tactile changes and may introduce unreliable predictions. The best performance is achieved with a window size of 12, which offers a good balance between informative future context and prediction reliability.

\begin{figure}[t]
\vspace{-1mm}
    \centering
    \includegraphics[width=\linewidth]{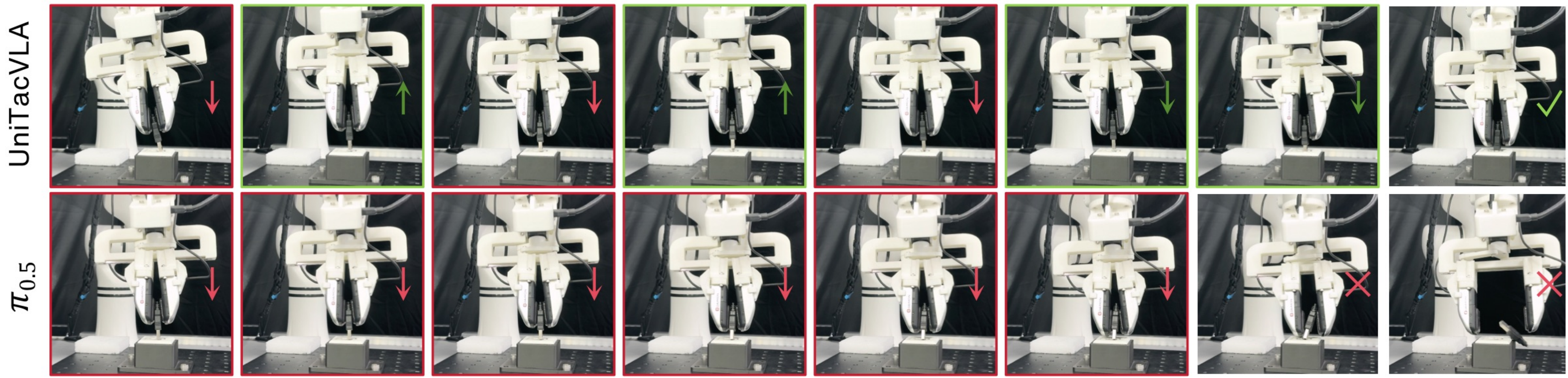}
    \vspace{-3mm}
    \caption{Analysis of the high-frequency controller on the USB insertion task. The controller produces timely residual corrections when transient collisions occur, improving task success rates.}
    \vspace{-4mm}
    \label{fig:controller}
\end{figure}

%
\textbf{3) Action-tactile mixed controller.}
The action-tactile mixed controller further improves both task success rate and execution stability. As shown in Table~\ref{tab:ablation}, introducing the controller increases the success rate from 52\% to 62\%, indicating that high-frequency tactile feedback provides additional benefits beyond tactile reasoning and prediction. This improvement is particularly important for dynamic contact tasks such as insertion and assembly. As illustrated in Fig.~\ref{fig:controller}, without the controller, the policy tends to execute actions in a more open-loop manner and struggles to recover from collisions or contact deviations in time. In contrast, the proposed controller refines actions online according to the predicted and observed tactile states, leading to more stable contact maintenance and more reliable recovery behavior~\textbf{(Q4)}.

\vspace{-2mm}
\section{Conclusion}
\label{sec:conclusion}
\vspace{-3mm}
In this paper, we presented UniTacVLA, a unified tactile learning framework that enhances VLA models for contact-rich robotic manipulation by modeling tactile signals as proactive interaction cues rather than passive auxiliary inputs. Through tactile chain-of-thought reasoning, coarse-to-fine future tactile prediction, and tactile-guided action correction, UniTacVLA jointly captures current contact states and future tactile dynamics to support more responsive manipulation. Real-world experiments demonstrate that our method consistently improves success rate, precision, and robustness across diverse contact-rich tasks, highlighting the importance of integrating tactile understanding and prediction for physically grounded robotic control.

\textbf{Limitations.}
Despite the promising results, our method has several limitations. First, the tactile data collected from teleoperated demonstrations may contain operator-dependent noise, which can make fine-grained tactile dynamics harder to learn. Second, although our multimodal fusion benefits tactile understanding and prediction, its robustness under severe visual occlusion, or incomplete language instructions remains insufficiently explored. Third, our current framework does not explicitly model force/torque signals, which may further improve performance in contact-rich manipulation.


\clearpage
\bibliography{example}  




\newpage
\appendix

\section{Dataset and Task Details}
\label{sec:appendix:dataset_task}

\subsection{Hardware Details}
\label{sec:appendix:hardware_details}

\begin{figure*}[h]
   \centering
    \includegraphics[width=0.6\linewidth]{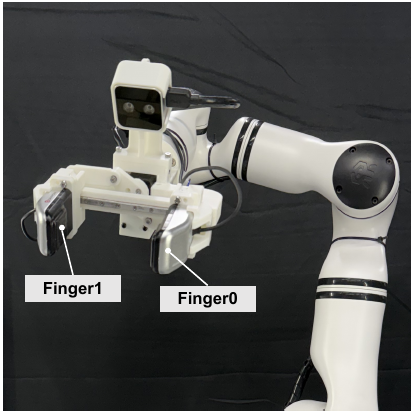}
    \caption{Gripper setup with two DM-Tac W visuo-tactile sensors mounted on the fingertips. The two tactile sensors are denoted as \textit{Finger 0} and \textit{Finger 1}.}
    \label{fig:gripper_setup}
\end{figure*}

All demonstrations are collected using a teleoperation system based on an ALOHA-style master-slave configuration. Both the master and slave manipulators use the same RealMan RM75B robotic arm. The slave robot is equipped with a 3D-printed parallel gripper, with two DM-Tac W visuo-tactile sensors mounted on the fingertips. Each tactile sensor records three-channel surface deformation signals, including depth deformation and shear deformation along the $x$- and $y$-axes, denoted as $[\mathrm{depth}, \mathrm{shear}_x, \mathrm{shear}_y]$. In this paper, the two tactile sensors are referred to as \textit{Finger 0} and \textit{Finger 1}, as shown in Figure~\ref{fig:gripper_setup}. Visual observations are captured by an Intel RealSense D405 wrist camera and an Intel RealSense L515 first-person-view camera.

\subsection{Task Details}
\label{sec:appendix:task_details}

\begin{figure*}[t]
    \centering
    \includegraphics[width=\textwidth]{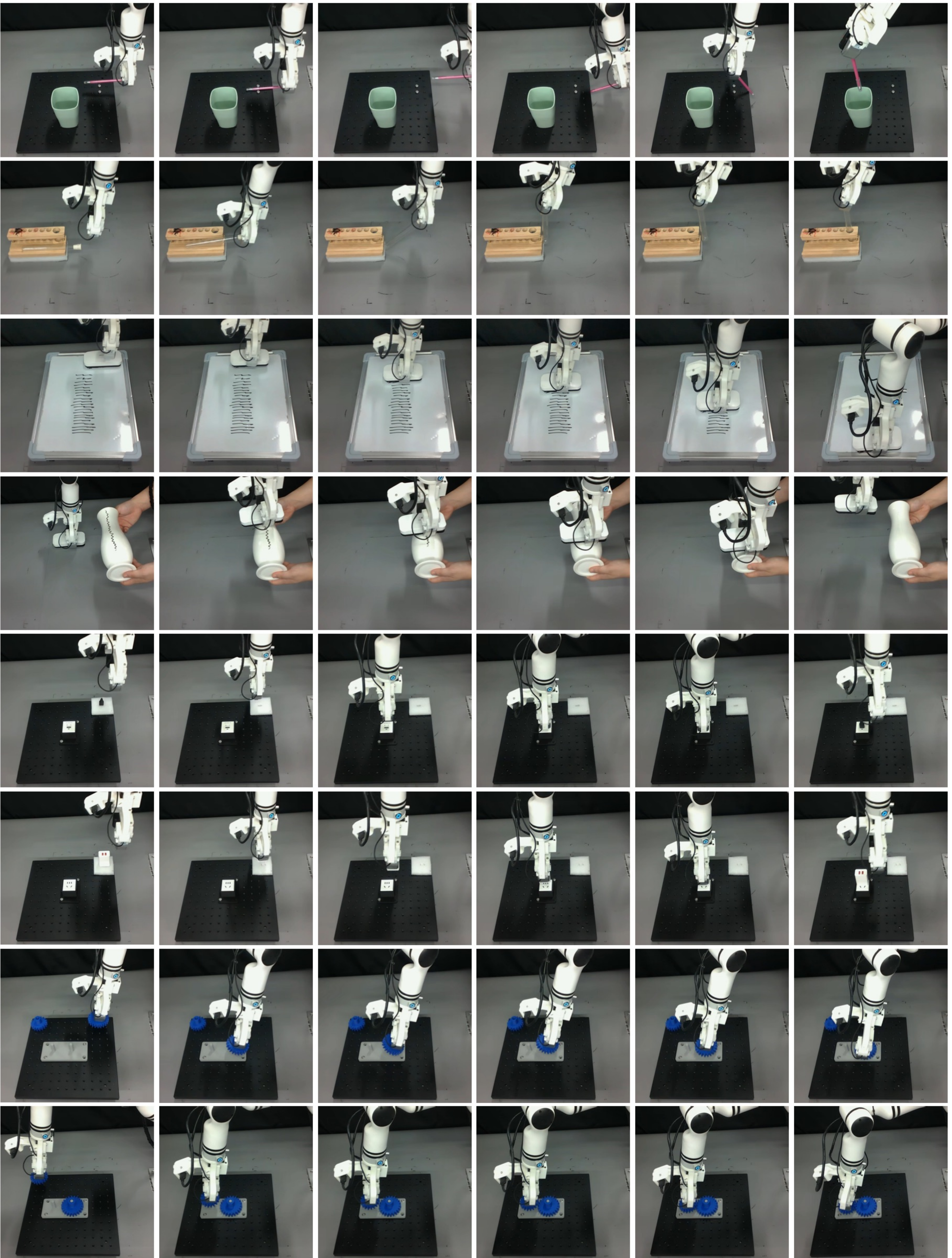}
    \caption{Representative trajectories and task setups for the four categories of contact-rich manipulation tasks: pick-and-adjust, pick-and-wipe, pick-and-insert, and pick-and-assemble.}
    \label{fig:task_setup}
\end{figure*}

We consider four categories of contact-rich manipulation tasks: pick-and-adjust, pick-and-wipe, pick-and-insert, and pick-and-assemble. Representative trajectories of each task category are shown in Figure~\ref{fig:task_setup}. The eight subtasks are defined as follows.

\begin{itemize} [leftmargin=*]
    \item \textbf{Pencil:} pick up the pencil, adjust its orientation, and place it into the cup.
    \item \textbf{Tube:} pick up the tube, adjust its angle, and place it into the tube rack.
    \item \textbf{Board:} pick up the whiteboard eraser and clean the whiteboard.
    \item \textbf{Vase:} pick up the board eraser and wipe the vase surface clean.
    \item \textbf{USB:} pick up the USB drive and insert it into the USB port.
    \item \textbf{Plug:} pick up the two-prong plug and insert it into the socket.
    \item \textbf{Gear-large:} pick up the large gear and insert it onto the corresponding gear shaft.
    \item \textbf{Gear-small:} pick up the small gear and insert it onto the corresponding gear shaft.
\end{itemize}

\subsection{Collection Details}
\label{sec:appendix:collection_details}

All demonstrations are collected through teleoperation at 30~Hz. For each subtask, we collect multiple trajectories, with approximately one hour of demonstration data per task on average.

During data collection, we collect two types of demonstration trajectories. The first type consists of normal expert demonstrations, which account for 80\% of the dataset. For all four task categories, the target object is placed on the tabletop during normal data collection. For the adjust tasks, we collect trajectories in which the object is adjusted from different initial orientations until it reaches a vertically downward pose. For the wipe tasks, we collect complete wiping trajectories with the target surface placed at different heights. For the insert tasks, we collect complete insertion trajectories with the slot initialized at different orientations. For the assemble tasks, we collect complete assembly trajectories with the gear shaft initialized at different orientations.

The second type consists of disturbance-recovery demonstrations, which account for the remaining 20\% of the dataset. During these demonstrations, human-induced perturbations are introduced and the operator provides corresponding corrective behaviors. Except for the wipe tasks, where the target object is manually supported by a human hand, the remaining three task categories are collected with the object placed on the tabletop. Specifically, in the adjust tasks, perturbations are introduced by changing the object placement angle. In the wipe tasks, perturbations are introduced by moving the target object downward. In the insert tasks, perturbations are introduced by changing the horizontal orientation of the slot during insertion. In the assemble tasks, perturbations are introduced by changing the orientation of the gear shaft.

In addition, to provide intuitive contact-intensity feedback to the demonstrator during data collection, we visualize the tactile signal strength in real time. Specifically, we quantify tactile intensity using the variance of the tactile signals and display its temporal variation as real-time feedback. This tactile feedback helps the demonstrator better perceive contact conditions during teleoperation, thereby improving data quality and further enhancing training stability and policy robustness.


\section{Implementation Details}

\subsection{System Overview}

UniTacVLA is built upon the $\pi0.5$ VLA backbone. During training, visual observations, language instructions, and tactile observations are processed by the multimodal backbones like SigLIP and VMAE. We introduce a set of learnable unified tactile queries to extract task-relevant tactile information from multimodal observations. Unless otherwise specified, the tactile prediction horizon is set to $K=12$. Tactile observations are encoded into compact latent representations using a variational masked autoencoder (VMAE). The policy operates directly on tactile latent representations during both training and inference. The tactile decoder is only used for visualization purposes and is not involved in policy execution. All experiments are trained on 4 NVIDIA A100 GPUs with a global batch size of 64 and deployed on a single RTX 5090 GPU for real-world evaluation.

\subsection{Model Architecture Details}

Table~\ref{tab:model_architecture} summarizes the architecture of each component in UniTacVLA.

\begin{table}[t]
\centering
\small
\begin{tabular}{lll}
\toprule
Module & Attribute & Value \\
\midrule

\multirow{3}{*}{Backbone}
& VLA Backbone & $\pi0.5$ \\
& Prediction Horizon & $K=12$ \\
& Unified Tactile Queries & $16\times K$, dim=2048 \\
\midrule

\multirow{5}{*}{Tactile Encoder}
& Architecture & VMAE (ViT-Small) \\
& Mask Ratio & 25\% \\
& Layers & 12 \\
& Attention Heads & 8 \\
& Hidden Dimension & 192 \\
& Latent Dimension & 32 \\
\midrule

\multirow{1}{*}{T-CoT Decoder}
& Language Model & Qwen3-0.6B \\
\midrule

\multirow{2}{*}{Coarse Predictor}
& Architecture & MLP \\
& Dimensions & $2048 \rightarrow 512 \rightarrow 32$ \\
\midrule

\multirow{3}{*}{Fine Predictor}
& Architecture & DiT \\
& Depth & 8 \\
& Width / Heads & 512 / 8 \\
\midrule

\multirow{3}{*}{Controller}
& Architecture & Transformer \\
& Layers & 6 \\
& Width / Heads & 192 / 3 \\
\bottomrule
\end{tabular}
\caption{Architecture details of UniTacVLA.}
\label{tab:model_architecture}
\end{table}

\paragraph{Unified tactile queries.}
We introduce $16\times K$ learnable unified tactile queries, where $K$ denotes the tactile prediction horizon. Each tactile query has the same hidden dimension as the backbone VLM token embedding (2048). These queries serve as a bridge between tactile understanding, tactile prediction, and action generation.

\paragraph{Tactile encoder.}
The tactile encoder is implemented using a variational masked autoencoder (VMAE). Specifically, we adopt a ViT-Small architecture consisting of 12 Transformer layers with 8 attention heads and hidden dimension 192. The encoded tactile representation is projected into a 32-dimensional latent space. During tactile representation learning, 25\% of tactile patches are randomly masked.

\paragraph{Tactile chain-of-thought decoder.}
To provide semantic supervision for tactile understanding, we employ Qwen3-0.6B as the language decoder. Unified tactile queries are used as conditioning tokens to autoregressively generate tactile chain-of-thought descriptions.

\paragraph{Coarse tactile predictor.}
The coarse tactile predictor is implemented as a two-layer multilayer perceptron:
\begin{equation}
z_t^{\text{coarse}}
=
\mathrm{MLP}(z_t^q),
\end{equation}
where the hidden dimensions are
\[
2048 \rightarrow 512 \rightarrow 32,
\]
with GELU activation.

\paragraph{Fine tactile predictor.}
The fine tactile predictor is implemented using a Diffusion Transformer (DiT) with 8 Transformer blocks, hidden dimension 512, and 8 attention heads. Instead of directly predicting future tactile latents, the DiT learns a conditional velocity field using flow matching.

Given the coarse tactile prediction $z_t^{\mathrm{coarse}}$ and the unified tactile queries $z_t^q$, we first construct an interpolated latent state

\begin{equation}
x_\tau
=
(1-\tau) z_t^{\mathrm{coarse}}
+
\tau z_t^{\mathrm{target}},
\quad
\tau \sim \mathcal{U}(0,1),
\end{equation}

where $z_t^{\mathrm{target}}$ denotes the ground-truth future tactile latent. The DiT predicts the corresponding velocity field

\begin{equation}
v_\theta
=
\mathrm{DiT}
\left(
x_\tau,
\tau,
z_t^q
\right).
\end{equation}

The model is optimized using the flow-matching objective

\begin{equation}
\mathcal{L}_{\mathrm{fine}}^{\mathrm{FM}}
=
\mathbb{E}_{\tau}
\left[
\left\|
v_\theta(x_\tau,\tau,z_t^q)
-
\left(
z_t^{\mathrm{target}}
-
z_t^{\mathrm{coarse}}
\right)
\right\|_2^2
\right].
\end{equation}

During inference, the DiT progressively transforms the coarse tactile latent into a refined future tactile latent conditioned on the unified tactile queries, producing tactile predictions with richer local contact details.

\textbf{Action-tactile mixed controller.}
The high-frequency action-tactile mixed controller is implemented as a lightweight Transformer with 6 layers, hidden dimension 192, and 3 attention heads. Given the future tactile latent sequence predicted by the tactile world model, we pair each predicted tactile latent with its corresponding policy action to construct a unified action-tactile chunk. This chunk, together with high-frequency robot state updates and real-time tactile latent observations, is fed into the action-tactile mixed controller. The robot state anchors the current physical configuration, while the policy action serves as the reference action to be refined. We further introduce a learnable \(\Delta\)action query token, which attends to the multimodal token sequence and predicts the residual action correction under the current contact condition. To prevent excessive deviation from the backbone policy, the residual correction is modulated by a gating network before being added to the original policy action for execution. 

\subsection{Training Details}

\begin{figure}[h]
    \centering
    \includegraphics[width=0.4\linewidth]{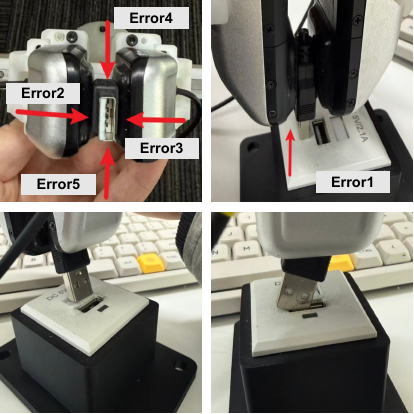}
    \caption{Tactile data collection for encoder pretraining using a handheld gripper with the same tactile sensing configuration as the robot gripper. The USB insertion task is shown as an example.}
    \label{fig:encoder_train}
\end{figure}

\textbf{Tactile representation learning.}
The tactile encoder is first pretrained using the variational masked autoencoding objective described in Eq.~(9), with a masking ratio of 25\%. The pretraining data consist of two sources. The first source is manually collected tactile data using a handheld gripper with the same tactile sensing configuration as the robot gripper. These data are organized into eight categories according to the corresponding manipulation tasks. Taking the USB insertion task as an example, the collection procedure is illustrated in Figure~\ref{fig:encoder_train}. The second source is extracted from teleoperation demonstrations, where frames with meaningful tactile responses are selected as additional tactile training samples.

Raw tactile images are stored as 16-bit sensor outputs and loaded with their original precision. During pretraining, the images are normalized from the uint16 range $[0, 65535]$ to $[0, 1]$, and then linearly rescaled to $[-1, 1]$ as the encoder input. The encoder compresses the normalized tactile observation into a compact latent representation $z \in \mathbb{R}^{32 \times 8 \times 8}$. Unlike the input and reconstructed image, the latent representation $z$ is not constrained to a fixed numerical range, but is learned through the variational masked autoencoding objective and depends on the training distribution. The decoder then reconstructs tactile observations from $z$ in the normalized image space, using a $\tanh$ output layer to produce values in $[-1, 1]$. For visualization or storage, the reconstructed outputs are rescaled back to $[0, 1]$ or converted to the uint16 range.

\begin{figure}[]
    \centering
    \includegraphics[width=0.4\linewidth]{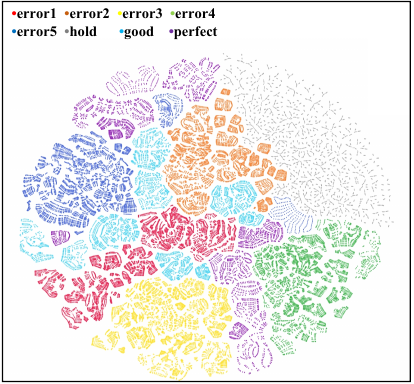}
    \caption{t-SNE visualization of the learned tactile latent space after encoder pretraining. The pretrained encoder produces discriminative representations that separate normal contact states from error cases.}
    \label{fig:encoder_tsne}
\end{figure}

After pretraining, we further analyze the learned tactile latent space using t-SNE. As shown in Figure~\ref{fig:encoder_tsne}, the pretrained encoder produces discriminative tactile representations and can effectively separate normal contact states from error cases, indicating that the learned latent space captures meaningful contact patterns for subsequent tactile reasoning and prediction.

\paragraph{Stage1 training.}
In the first stage, all components except the action-tactile mixed controller are jointly optimized on clean expert demonstrations. 
This includes the VLM backbone, unified tactile queries, tactile encoder, tactile chain-of-thought module, and coarse-to-fine tactile predictor. The optimization objective is

\begin{equation}
L_{\text{stage1}}
=
L_{\text{action}}^{\text{FM}}
+
\lambda_{\text{sem}}L_{\text{semantic}}^{\text{CE}}
+
\lambda_{\text{coarse}}L_{\text{coarse}}^{1}
+
\lambda_{\text{fine}}L_{\text{fine}}^{\text{FM}},
\end{equation}

where

\begin{equation}
\lambda_{\text{sem}}
=
\lambda_{\text{coarse}}
=
\lambda_{\text{fine}}
=
0.1.
\end{equation}

\paragraph{Stage2 training.}
After stage-1 convergence, all previously trained modules are frozen. 
Only the action-tactile mixed controller is optimized using the residual action correction objective. 
The controller is trained on a mixture of clean expert demonstrations and disturbance-recovery trajectories collected under human interventions. The optimization objective is
\begin{equation}
\mathcal{L}_{\mathrm{stage2}}
=
\mathcal{L}_1^{\mathrm{ctrl}}.
\end{equation}

\paragraph{Optimization.}
All models are optimized using AdamW. The optimization hyperparameters of stage1 and stage2 are summarized in Table~\ref{tab:training_hyper}.

\begin{table}[h]
\centering
\begin{tabular}{lc}
\toprule
Hyperparameter & Value \\
\midrule
Optimizer & AdamW \\
Learning rate & $2.5\times10^{-5}$ \\
Weight decay & 0.01 \\
Gradient clipping & 1.0 \\
Learning rate schedule & Cosine decay \\
Warmup steps & 1000 \\
Global batch size & 64 \\
Training GPUs & 4$\times$A100 \\
Training steps & 5k--10k \\
\bottomrule
\end{tabular}
\caption{Training hyperparameters of stage1 and stage2.}
\label{tab:training_hyper}
\end{table}


\paragraph{Training schedule.}
Depending on the dataset size, UniTacVLA is trained for approximately 5k--10k optimization steps, corresponding to roughly three training epochs. All models are trained on four NVIDIA A100 GPUs and evaluated on a single NVIDIA RTX 5090 GPU.

\subsection{Tactile Chain-of-Thought Generation Details}

To provide semantic supervision for tactile understanding, we design a structured tactile chain-of-thought (T-CoT) template. Rather than directly describing tactile observations, T-CoT decomposes tactile reasoning into three sequential components:

\begin{equation}
\mathrm{TCoT}_t
=
\{s_t, m_t, a_t\},
\end{equation}

where $s_t$ denotes interaction-stage reasoning, $m_t$ denotes modality-dependency reasoning, and $a_t$ denotes action-guidance reasoning.

The T-CoT annotations are generated according to the following prompt template:

\begin{figure}[t]
\centering
\fbox{
\parbox{0.95\linewidth}{
\small
\textbf{T-CoT Prompt Template}

\vspace{0.5em}

\textbf{[Interaction Stage Analysis]}

Determine the current interaction stage:

\begin{itemize}
\item Loose: physical interaction has not yet occurred.
\item Holding: physical interaction is present and relatively stable.
\item Contact: contact conditions change rapidly and may involve collision, slip, insertion, or alignment.
\end{itemize}

Describe the current interaction stage and explain the corresponding physical state.

\vspace{0.5em}

\textbf{[Modality Dependency Analysis]}

Analyze:

\begin{itemize}
\item contact intensity,
\item force magnitude,
\item slip risk,
\item visual observability,
\item tactile signal variation.
\end{itemize}

Determine whether vision or tactile sensing is more reliable under the current interaction condition.

\vspace{0.5em}

\textbf{[Action Guidance Analysis]}

Based on the interaction stage and modality dependency:

\begin{itemize}
\item explain the current manipulation status,
\item identify potential risks,
\item describe which modality should dominate decision making,
\item provide action guidance for the policy.
\end{itemize}
}
}
\caption{Prompt template used for tactile chain-of-thought annotation.}
\label{fig:tcot_prompt}
\end{figure}

Figure~\ref{fig:tcot_prompt} defines the structured reasoning process used to generate tactile chain-of-thought annotations. The resulting reasoning chain serves as semantic supervision for learning the unified tactile representation.

An example generated T-CoT is shown below:

\begin{lstlisting}
The robot is in the contact stage. Physical interaction transforms sharply. Vision may be partially occluded. Tactile information is rich. Tactile feedback changes sharply. Slip risk is high. The robot is physically interacting with the object. Tactile information becomes more reliable than visual observations. The controller should focus on contact stabilization and fine-grained corrective actions.
\end{lstlisting}

The language decoder is only used during training for semantic supervision. During inference, the decoder is discarded and only the learned tactile representations are retained for policy execution.


\section{Additional Experimental Results}
\label{sec:appendix:additional_results}

In this section, we provide additional results to further analyze the behavior of UniTacVLA. 
We focus on three aspects: tactile prediction across different manipulation tasks, state-aware tactile reasoning, and the behavior of the action-tactile mixed controller. 

\subsection{Tactile Prediction}
\label{sec:appendix:tactile_prediction}

\begin{figure*}[t]
    \centering
    \includegraphics[width=\linewidth]{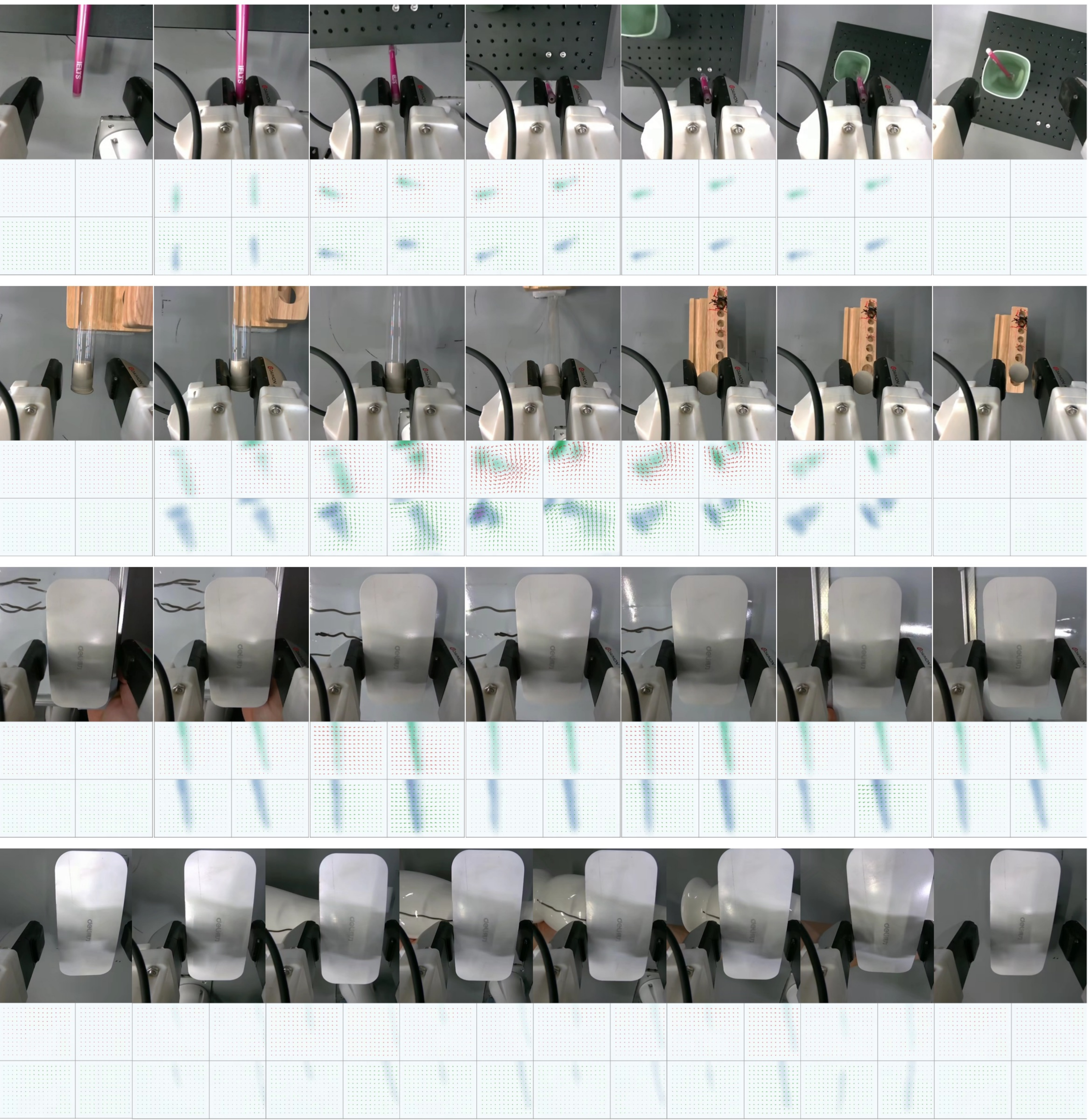}
    \caption{
    Qualitative comparison between fine-level tactile prediction and ground-truth tactile observations on adjustment and wiping tasks. 
    Red arrows indicate predicted tactile changes, while green arrows indicate the corresponding ground-truth tactile changes.
    }
    \label{fig:tacpretask1}
\end{figure*}

\begin{figure*}[t]
    \centering
    \includegraphics[width=\linewidth]{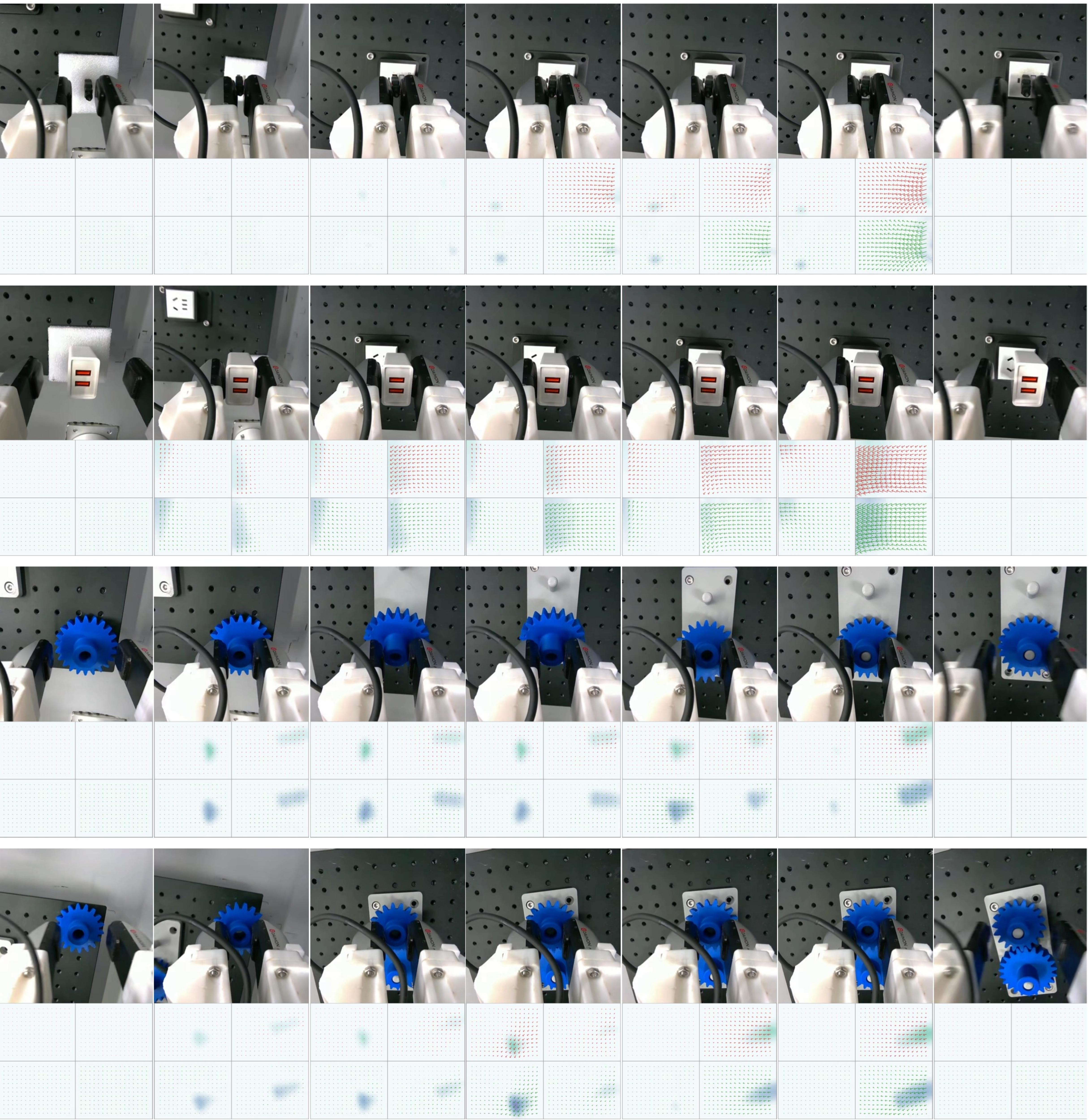}
    \caption{
    Qualitative comparison between fine-level tactile prediction and ground-truth tactile observations on insertion and assembly tasks. 
    Red arrows indicate predicted tactile changes, while green arrows indicate the corresponding ground-truth tactile changes.
    }
    \label{fig:tacpretask2}
\end{figure*}

To further evaluate the quality of future tactile prediction, we visualize the fine-level tactile predictions and the corresponding ground-truth tactile observations across different task categories. 
As shown in Figure~\ref{fig:tacpretask1} and Figure~\ref{fig:tacpretask2}, UniTacVLA can predict the major spatial deformation patterns and temporal contact variations under diverse contact-rich manipulation scenarios.
In particular, for adjustment and wiping tasks, the predicted tactile signals capture the transition from free-space motion to stable contact, as well as the local deformation caused by object interaction or surface contact. 
These results indicate that the learned tactile latent space preserves physically meaningful contact dynamics rather than only reconstructing static tactile patterns.

\begin{figure*}
    \centering
    \includegraphics[width=\linewidth]{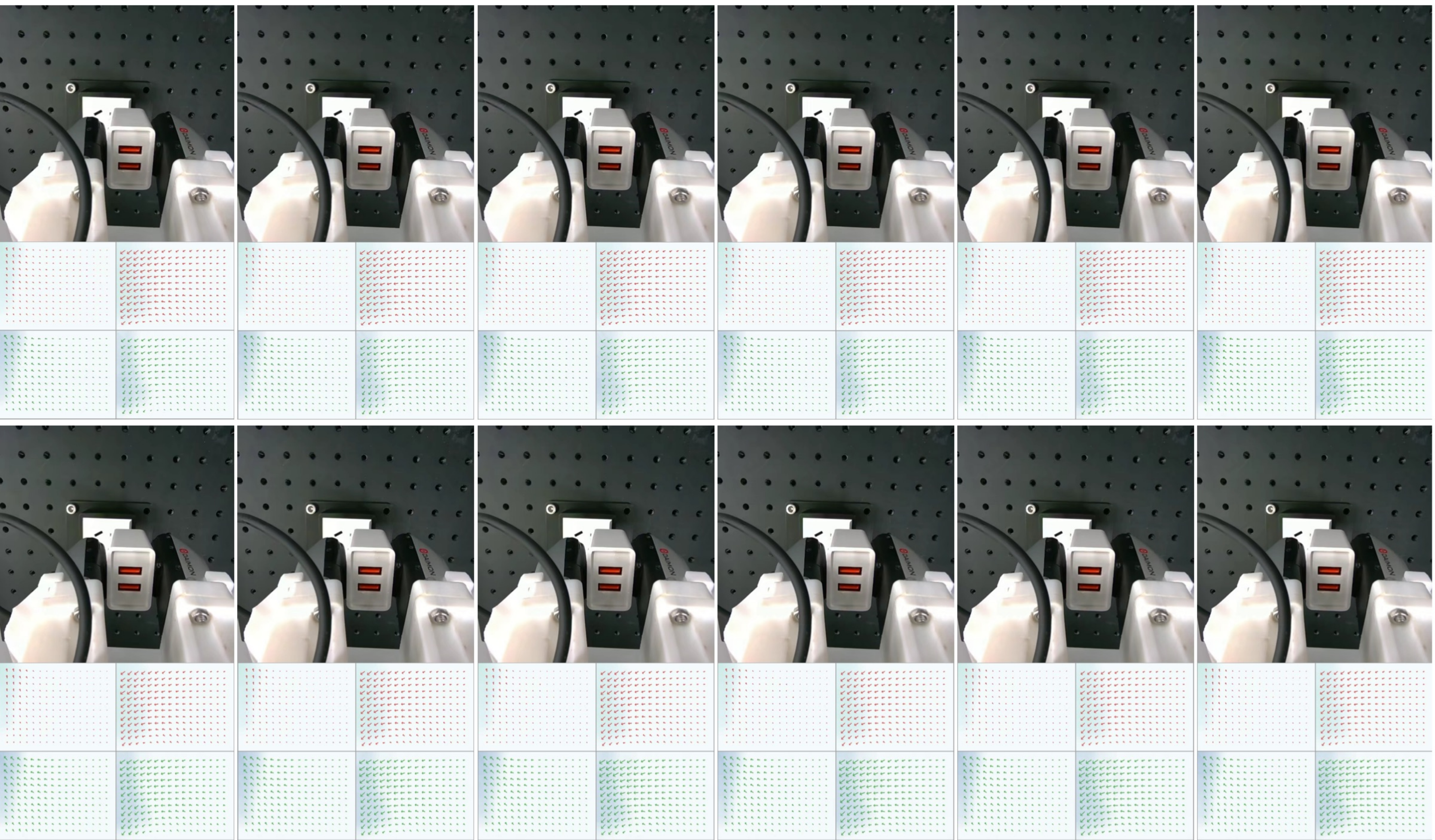}
    \caption{
    Qualitative comparison of tactile prediction within an action-tactile pair chunk on the plug insertion task. 
    Red arrows indicate predicted tactile changes, while green arrows indicate the corresponding ground-truth tactile changes.
    }
    \label{fig:tacprechunk}
\end{figure*}

We further visualize the predicted tactile sequence within an action-tactile pair chunk using the plug insertion task as an example. 
As shown in Figure~\ref{fig:tacprechunk}, the predicted tactile sequence follows the evolution of the real tactile observations during the action chunk. 
When the plug approaches the socket and contact becomes stronger, the predicted tactile deformation also becomes more concentrated and structured. 
This suggests that the coarse-to-fine tactile predictor can provide future-aware contact cues for action generation and online correction.

\subsection{T-CoT State Awareness}
\label{sec:appendix:tcot_state_awareness}

\begin{figure}
    \centering
    \includegraphics[width=\linewidth]{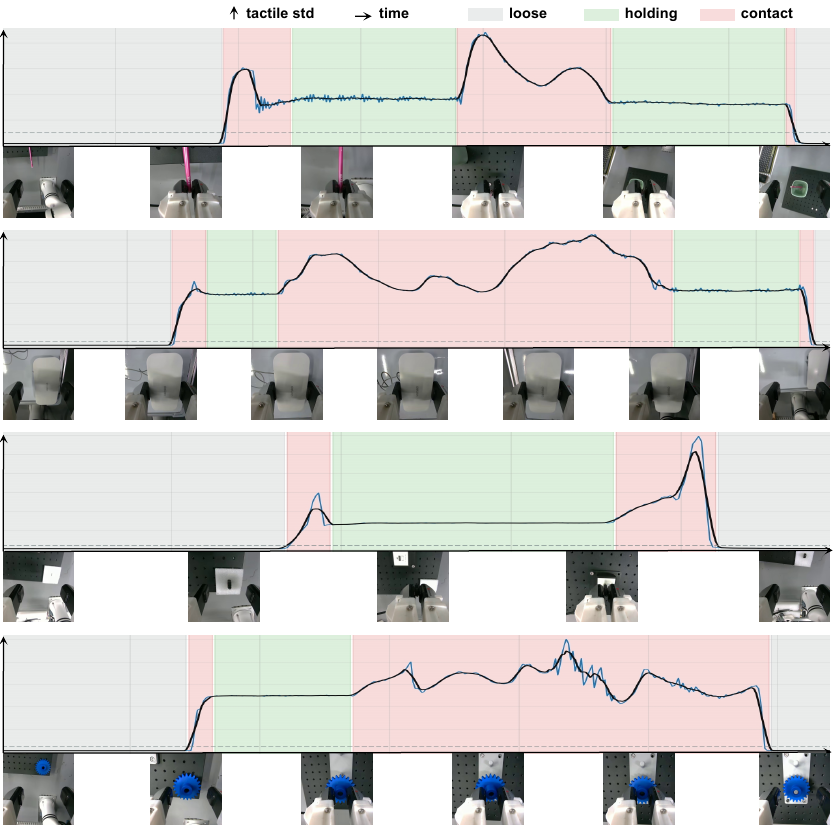}
\caption{
Visualization of the state-awareness capability of T-CoT across different tasks.
The vertical axis denotes the average standard deviation of the depth and shear components from real tactile observations.
T-CoT can distinguish different contact modes from the current latent tactile state, and its state understanding is consistent with the variation trend of real tactile signals.
These results show that T-CoT can correctly recognize loose, holding, and contact states, demonstrating its ability to capture task-relevant contact semantics.
}
    \label{fig:tcot_state_awareness}
\end{figure}

In this section, we demonstrate the state-awareness capability of T-CoT.
As shown in Figure~\ref{fig:tcot_state_awareness}, we visualize the state
differentiation results produced by T-CoT across different tasks. These results
show that T-CoT can accurately perceive the active contact mode from the current
latent tactile state, thereby providing task-stage-aware guidance for downstream
action generation.

\subsection{Action-Tactile Mixed Controller}
\label{sec:appendix:act_tac_controller}

The action-tactile mixed controller provides high-frequency closed-loop corrections on top of the low-frequency action chunks generated by the backbone policy. 
Instead of relying solely on the initially predicted action chunk, the controller continuously refines the executed action according to both the predicted future tactile states and the current tactile observations. 
This design allows the policy to combine proactive contact anticipation with reactive tactile feedback during execution.

\begin{figure}
    \centering
    \includegraphics[width=\linewidth]{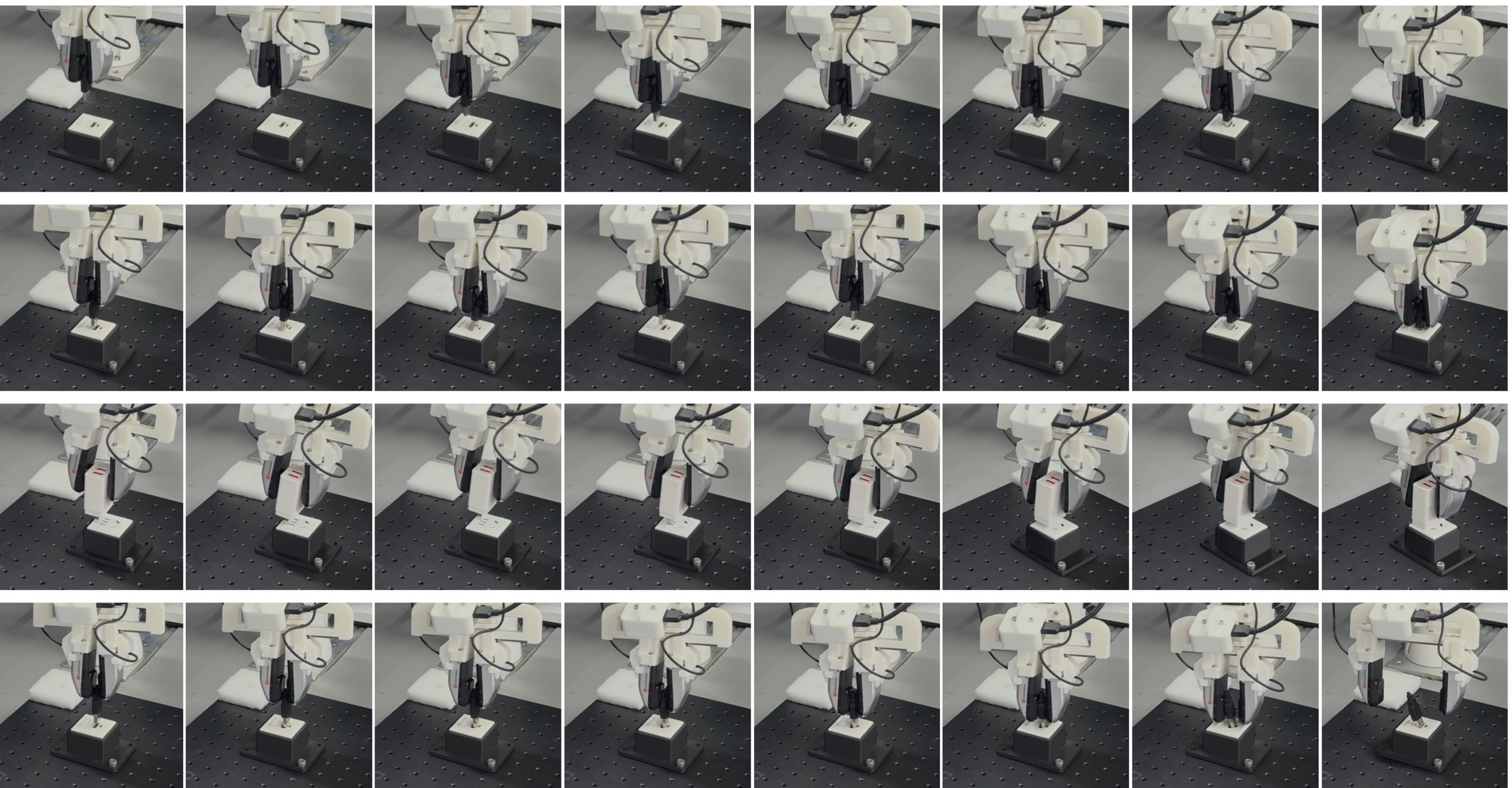}
\caption{
Qualitative behaviors of the action-tactile mixed controller in insertion tasks.
The first row shows proactive correction in USB insertion, while the second and third rows show error-correction behaviors in USB and plug insertion, respectively.
The controller enables the policy to recover from contact deviations and complete the insertion.
In contrast, the fourth row shows a USB insertion failure without the controller, where the policy fails to correct the deviation in time and the task collapses.
}
    \label{fig:controller_appendix}
\end{figure}

The effect of the controller is most evident in high-frequency dynamic contact tasks, where small contact deviations may quickly lead to collision, slipping, or jamming. 
When the predicted tactile state indicates an upcoming unstable contact, the controller can adjust the action before the error becomes visually obvious. 
When unexpected disturbances occur during execution, real-time tactile feedback further enables the controller to respond quickly and recover from contact deviations. 
As shown in Figure~\ref{fig:controller_appendix}, in insertion tasks, the model with the controller produces timely corrective motions when contact errors occur, leading to successful recovery. 
In contrast, the model without the controller tends to follow the original action chunk in a more open-loop manner and fails to correct the deviation in time, resulting in insertion failure.

\end{document}